\documentclass{article}

\usepackage{microtype}
\usepackage{graphicx}
\usepackage{booktabs} % for professional tables

\usepackage{amsmath}
\usepackage{pgf-pie}
\usepackage{listings}
\usepackage{amssymb} 
\usepackage{pgfplots}
\usepackage{caption}
\usepackage{subcaption}
\usepackage{pifont}% http://ctan.org/pkg/pifont
\newcommand{\cmark}{\ding{51}}%
\newcommand{\xmark}{\ding{55}}%
\newcommand{\onemark}{\ding{172}}
\newcommand{\twomark}{\ding{173}}
\newcommand{\threemark}{\ding{174}}

\usepackage{hyperref}

\usepackage[accepted]{icml2021}

\icmltitlerunning{EL-Attention: Memory Efficient Lossless Attention for Generation}

\begin{document}

\twocolumn[
\icmltitle{EL-Attention: Memory Efficient Lossless Attention for Generation}



\icmlsetsymbol{equal}{*}

\begin{icmlauthorlist}
\icmlauthor{Yu Yan}{ms}
\icmlauthor{Jiusheng Chen}{ms}
\icmlauthor{Weizhen Qi}{equal,ustc}
\icmlauthor{Nikhil Bhendawade}{equal,ms}
\icmlauthor{Yeyun Gong}{equal,msra}
\icmlauthor{Nan Duan}{msra}
\icmlauthor{Ruofei Zhang}{msca}
\end{icmlauthorlist}

\icmlaffiliation{ms}{Microsoft, Redmond, WA, USA}
\icmlaffiliation{msra}{Microsoft Research Asia}
\icmlaffiliation{ustc}{University of Science and Technology of China}
\icmlaffiliation{msca}{Microsoft, Sunnyvale, CA, USA}

\icmlcorrespondingauthor{Yu Yan}{yyua@microsoft.com}

\icmlkeywords{Attention, Transformer, Inference Speed, Generation Speed}

\vskip 0.3in
]



\printAffiliationsAndNotice{\icmlEqualContribution} % otherwise use the standard text.

\begin{abstract}
Transformer model with multi-head attention requires caching intermediate results for efficient inference in generation tasks. 
However, cache brings new memory-related costs and prevents leveraging larger batch size for faster speed.
We propose memory-efficient lossless attention (called EL-attention) to address this issue. It avoids heavy operations for building multi-head keys and values, cache for them is not needed.
EL-attention constructs an ensemble of attention results by expanding query while keeping key and value shared. It produces the same result as multi-head attention with less GPU memory and faster inference speed. 
We conduct extensive experiments on Transformer, BART, and GPT-2 for summarization and question generation tasks. The results show EL-attention speeds up existing models by 1.6x to 5.3x without accuracy loss.
\end{abstract}

\section{Introduction}
Transformer model with multi-head attention achieves success in various generation tasks, such as text generation~\citep{raffel2019exploring, radford2019language, lewis-etal-2020-bart, brown2020language}, image generation~\citep{parmar2018image, cho-etal-2020-x}, and music generation~\citep{huang2018music}. However, inference speed is a serious problem in generation models. Recently, a variety of methods have been proposed for the speed up of Transformer and variant models. 
Many methods focus on reducing complexity on sequence length, like restricting tokens which can be looked at~\citep{NEURIPS2020_c8512d14,beltagy2020longformer}, using sort~\citep{tay2020sparse} or hash technology~\citep{Kitaev2020Reformer:}, keeping cumulative states~\citep{pmlr-v119-katharopoulos20a}, and compressing dimension~\citep{goyal2020power,wang2020linformer}. 
Others study reducing model size to accelerate inference by pruning layer~\citep{fan2019reducing} or training a smaller student model~\citep{shleifer2020pre}.
Another way is non-autoregressive generation~\citep{gu2018nonautoregressive,lee-etal-2018-deterministic,qi2020bang} which generates all tokens at once, instead of predicting the next token step by step.
While these excellent methods can effectively speed up the models, they require users to train a new model, and it is hard to apply them to an existing model directly. Moreover, most of them suffer more or less accuracy loss~\citep{tay2021long}. 

In this paper, we explore generation speedup by optimizing cache size and memory movement rather than reducing computational complexity. 
This choice is based on an insight from studying the generation process. Multi-head attention needs to convert query, key, and value to their multi-head format for assembling attention results.  
Under the context of incremental generation, a query is a feature vector belonging to $\mathbb{R}^{d_m}$, while key and value are sequences of $n$ feature vectors with shape $\mathbb{R}^{n\times d_m}$. 
Changes in key and value are minimal  during an incremental decoding step, and they are usually cached to avoid complex and duplicated computation. 
With cache, overall speed is much faster than without cache, but cache maintenance cost is still non-trivial. 
There is a significant room to achieve quicker speed by reducing memory complexity.

We present a new memory-efficient lossless attention, called \emph{EL-attention}. It can speed up inference by reducing cache size and memory movement complexity. 
Memory used for caching input related model states is reduced from $\mathcal{O}(Ld_m)$ to $\mathcal{O}(d_m)$ where $L$ is number of decoder layers and $d_m$ is model dimension. 
EL-attention further reduces memory movement when using beam search or other search methods to produce many candidates for each input. 
We achieve this by only expanding query when constructing an ensemble of attention results. Our method avoids converting key and value to multi-head format. All queries from different heads and beams share the same key and value. 
Despite the change of attention, our method can be compatible with multi-head attention via a particular converted form of query. 
By taking advantage of the saved memory movement, our method can accelerate inference under the same setting as the baseline.
Also, our method saves a massive amount of memory. This saved memory can be harvested in multiple ways, like fitting a bigger model in GPU, or using a larger batch size for further speed up.

To summarize our contributions:
\begin{enumerate}
  \item We propose a new attention method called EL-attention, which can replace multi-head attention at the inference stage to generate the same results with smaller cache size and less memory movement.
  \item We evaluate EL-attention on the Transformer, BART, GPT-2 model for summarization tasks (CNN/DailyMail and XSum) and question generation task (SQuAD 1.1).
  It shows 1.6x to 5.3x speedup across all models on these tasks in beam search, diverse beam search, and greedy search.
  \item We show that EL-attention uses 96x lesser memory for caching input related states in BART model, which supports running with 10x larger batch size and 5x speed.
  It is potentially valuable for memory limited devices, like mobile and IoT devices.
\end{enumerate}

\section{Background}
We first introduce Transformer~\cite{NIPS2017_3f5ee243} under generation context, then describe speed analysis.
 
\subsection{Scaled Dot-Product Attention}
Attention allows one position to grab information from other positions. 
Given key matrix and value matrix K, V $\in$ $\mathbb{R}^{n\times d}$, and a query vector Q $\in \mathbb{R}^{d}$, the attention is calculated as: 
\begin{equation}
\textrm{Att}(Q,K,V) = \textrm{softmax}(\frac{QK^T}{\sqrt{d}})V 
\label{equation.singlehead_attention} 
\end{equation}

\subsection{Multi-Head Attention}
In multi-head attention, independent attentions are applied in parallel to obtain the final result\footnote{This notation, equivalent to the ``concatenation” formulation from (Vaswani et al., 2017) is used to ease exposition in the following sections.}:
\begin{equation}
\begin{aligned}
&\textrm{MultiHead}(Q, K, V) = \sum_{i=1}^{h} \textrm{Att}(Q_i,K_i,V_i)W_i^O \\
&\textrm{where}\: Q_i=QW_i^Q, \: K_i=KW_i^K, \: V_i=VW_i^V 
\end{aligned}
\label{equation.multihead_attention} 
\end{equation}

Here, $W_i^Q$, $W_i^K$, $W_i^V$ $\in$ $\mathbb{R}^{d_m\times d_k}$ and $W_i^O$ $\in$ $\mathbb{R}^{d_k\times d_m}$. 
The $h$ is the number of heads, and $Q_i$, $K_i$, $V_i$ represent the query, key, and value for the i-th head respectively. 
The $d_k$ is typically set to $\frac{d_m}{h}$. 

\subsection{Incremental Decoding}
\begin{figure}[h]
    \centering
	\includegraphics[width = 3.5 in]{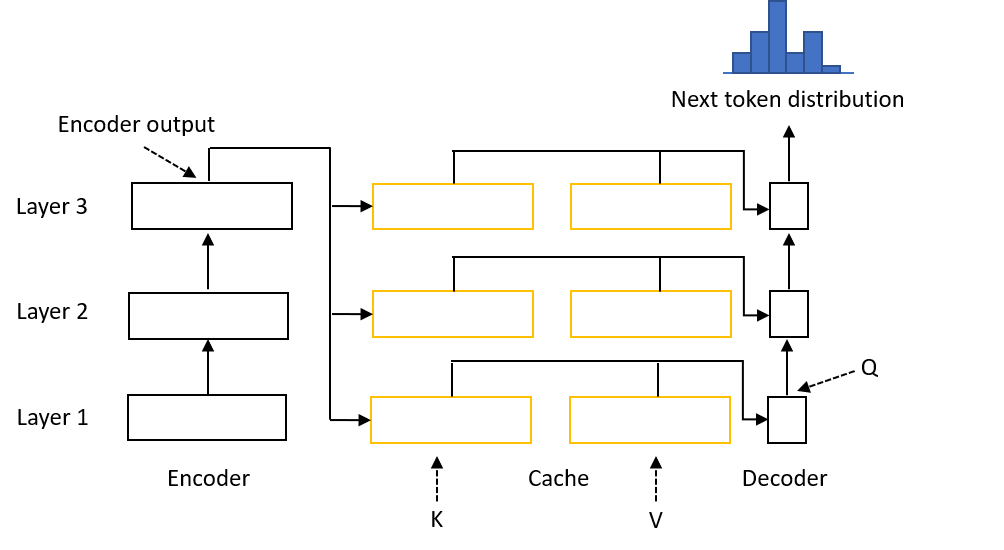}
	\caption{The cache used in encoder-decoder attention during incremental decoding.}
	\label{figure.incremental}
\end{figure}

For Transformer inference, prediction of the next token depends on the input sequence and previously generated sequence. The last token attends to its previous output and input at each step. States (key and value) are cached and re-used to avoid redundant calculation.  
See Figure~\ref{figure.incremental}, although each decoder layer attends to the same encoder output, it needs to build its own cache for storing key and value due to each layer having different weight parameters $W^K$ and $W^V$.

\subsection{Speed Analysis}
\label{subsecion.speed_limitation}
Arithmetic intensity (AI) is a ratio of floating-point operations performed to data movement (FLOPs/Bytes). Speed performance (GFLOP/s) is:
\begin{equation}
\textrm{GFLOP/s} = \textrm{min}
    \begin{cases}
      \textrm{Peak GFLOP/s} \\
      \textrm{Peak GB/s} \times \textrm{AI}
    \end{cases}
\label{equation.gflops} 
\end{equation}

A function is bounded by memory when AI is small, and its threshold ranges from ten to a hundred for V100 GPU on single and half precision~\cite{yang2020hierarchical,wang2020hierarchical}. 

Decoding is largely bounded by memory bandwidth due to low arithmetic intensity~\citep{shazeer2019fast,tay2020sparse}. Its speed is also affected by latency because of the small computational scale per instruction. 

\begin{figure*}	
	\centering
	\begin{subfigure}[t]{3in}
		\centering
		\includegraphics[width=3in]{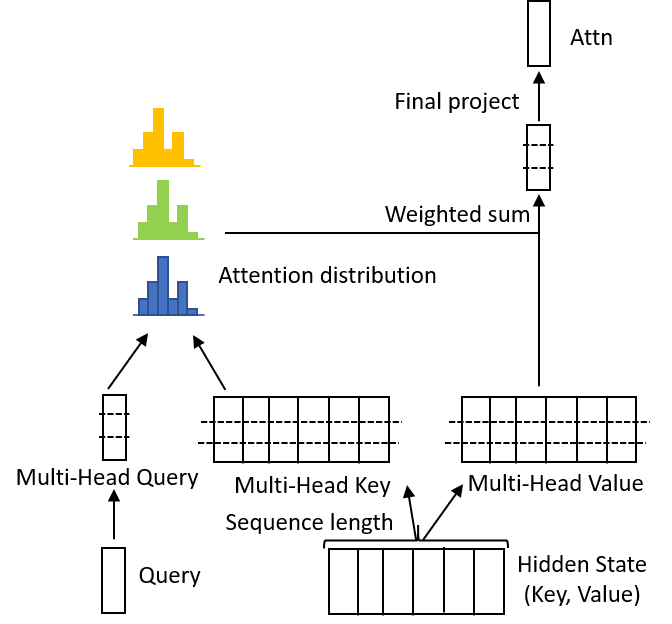}
		\caption{Multi-Head Attention}\label{figure.multi_head_attention}		
	\end{subfigure}
	\quad
	\begin{subfigure}[t]{3in}
		\centering
		\includegraphics[width=3in]{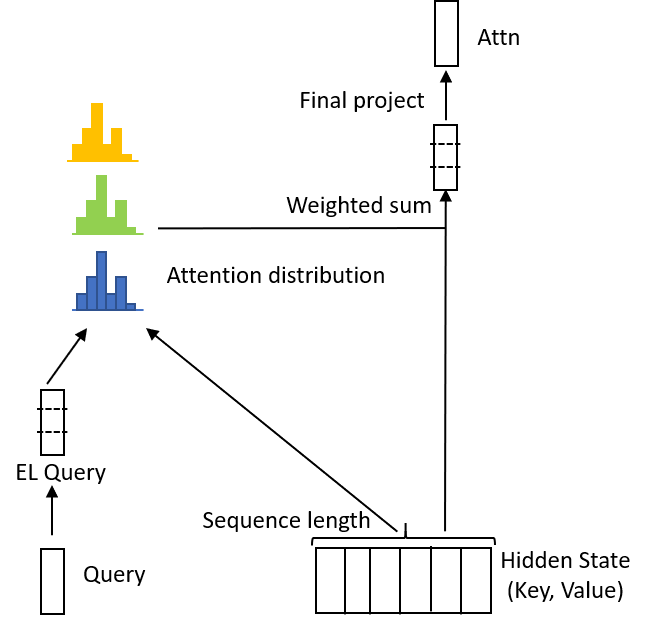}
		\caption{EL-Attention}\label{figure.el_attention}
	\end{subfigure}
	\caption{(a) Multi-head attention: query, key, and value are converted to their multi-head formats respectively, then the attention probability and attention result for each head are calculated; at last, results are aggregated to get the final output. (b) EL-attention: it only applies linear conversion to query. The hidden state, which is encoder output or previous layer output, is directly used as key and value without conversion.}
	\label{figure.main}
\end{figure*}

\section{Method}
\label{section.method}
In this section, we present memory-efficient lossless attention (EL-attention), which is designed for high inference speed and low memory requirement. 

We start by introducing EL-attention technical detail in \S~\ref{subsection.el_attention}. In \S~\ref{subsection.application}, we present how to use EL-attention to speed up existing vanilla Transformer-based models. 
Finally, we present theoretical speed analysis in \S~\ref{subsection.theoretical_analysis}.

\subsection{EL-Attention}
\label{subsection.el_attention}
Our method EL-attention constructs an ensemble of attention results by expanding query only and sharing key and value for all heads.
The overall workflow of EL-attention is illustrated in Figure~\ref{figure.el_attention}. 
We first convert query to multi-head format via expanding. Then it directly multiplies with hidden state, which is either encoder output or previous layer output, to calculate attention probability.
Intermediate attention result is computed by multiplying attention probability with the same hidden state.
Finally, a linear conversion is applied to produce the output. 
The multi-head attention (see Figure~\ref{figure.multi_head_attention}), which is used in the vanilla Transformer, has linear transformation for all query, key, and value to get their multi-head formats.

EL-attention introduces two feed-forward networks (FFN) for calculating an ensemble of attention results:
\begin{equation}
\textrm{EL}(Q, K, V) = \sum_{i=1}^{h} \textrm{FFN}_i^O(\textrm{Att}(\textrm{FFN}_i^Q(Q),K,V))) 
\label{equation.multihead_query_head} 
\end{equation}
Let FFN$_i^Q$ and FFN$_i^O$: $\mathbb{R}^{d_m}$ $\rightarrow$ $\mathbb{R}^{d_m}$ be conversion functions for query and individual attention results, the choice of this functions can be arbitrary, like  one or multiple linear conversions, or an activation function. 
In this work, the following forms are adopted to be completely compatible with vanilla transformer:
\begin{equation}
\begin{aligned}
&\textrm{EL-Q}_i=\textrm{FFN}_i^Q(Q)=QW_i^{Q}(W_i^{K})^T \\
&\textrm{and}\: \textrm{FFN}_i^O(X)=XW_i^{V}W_i^{O}
\end{aligned}
\label{equation.query_head} 
\end{equation}
where $W_i^Q$, $W_i^K$, $W_i^V$ $\in$ $\mathbb{R}^{d_m\times d_k}$ and $W_i^O$ $\in$ $\mathbb{R}^{d_k\times d_m}$, these weight parameters are same as the ones in multi-head attention. 
The normalization factor $d$ in attention softmax Equation~\ref{equation.singlehead_attention} is $d_k$ to keep same as multi-head attention, $d_k$ is the rank of EL-Q$_i$ while $d_m$ is its dimension. 

A critical difference between EL-attention and multi-head attention is how to assemble attention results. 
EL-attention can provide an ensemble result without projecting key and value to sub-spaces, only query is projected to low-rank format while keeping dimension size unchanged.
In multi-head attention, all query, key, and value are projected to sub-spaces, requiring high computational complexity or large memory size for caching.

\subsubsection{Efficient Implementation}
\label{subsubsection.efficient_implementation}
%To reduce the cache size and memory movement, several changes are made on top of Transformer implementation. Comparison with pseudocode are presented in Appendix B. 
Several changes are made on the generation implementation to reduce the cache size and memory movement. We present pseudocode to show the key difference between EL-attention and multi-head attention in Appendix A. 

First, as EL-attention does not depend on multi-head key and value, their cache can be entirely removed for all encoder-decoder attention sub-layers. Only encoder output is saved, and it is shared for all decoder layers. 
EL-attention reduces cache size by 2$L$ times for encoder-decoder attention, where $L$ is the number of layers in a decoder. 
In self-attention, we only store output from the previous layer, instead of both key and value, thereby reducing cache size by half. 

Second, since multiple query heads are mapped to the same key in EL-attention, we can reduce memory movement during attention calculation.
To recall batch matrix multiplication (BMM) background, BMM of two 3D matrices is equivalent to computing multiple 2D matrix multiplications in parallel. However, this good property requires two 3D matrices to have their first dimension size equal.
The first dimensions for query and key need to be adjusted to meet the requirement. 
Key generally uses a bigger memory size than query because sequence length for query is always one in incremental decoding. In contrast, key's length is either input length or previous output length. 
Repetitive loading of key and value linearly slows down the speed in this case. 
Our implementation involves reshaping query to match the first dimension of key. We decrease query's first dimension and increase its second dimension, and then one matrix multiplication can generate scores for all heads, which avoids loading key repetitively. 
%A sub-optimal implementation is to expand key's first dimension logically without physic data copy, but this cannot avoid memory loading duplication. 

In beam search or other searching methods, which generate many outputs per input, we can map multiple queries to one key and then the memory movement can be further reduced. 

\subsubsection{Exchangeable with Multi-Head Attention}
\label{subsubsection.exchangeable}
Our method EL-attention can provide the same functional ability as multi-head attention.
By leveraging the associative property of matrix multiplication, the multi-head attention formula can be derived from EL-attention, see the proof in Appendix B. 
We will present more generation result analyses in \S~\ref{subsection.quality}.

\subsection{EL-Attention Applications}
\label{subsection.application}
In this work, we explore using EL-attention to replace multi-head attention at the inference stage for speed up. 
We show how to use EL-attention in transformer model and language model. 

In text generation tasks, input length is generally longer than output length.
Query-input attention is much heavier than query-previous output attention in terms of computational and memory complexity because at each step query attends to the whole input but only attends to the previously generated output, which gradually grows from one to max output length.
With this in mind, we only apply EL-attention to calculate attention for input related part. 

\noindent\textbf{Encoder-Decoder Attention} For a model having encoder-decoder architecture, we apply EL-attention for encoder-decoder attention while keeping self-attention unchanged.

\noindent\textbf{Self-Attention} For a decoder-only model, input (prefix text) and output sequences are concatenated to feed into the decoder. Both input and previous output tokens are attended to in the self-attention calculation. In this case, we apply EL-attention to calculate attention for the input part, while multi-head attention is still used to handle the previous output part.
Attention weights for the input and the previous output are computed separately, then concatenated before applying softmax for computing attention probability, then split for building attention result from input and output, and followed by element-wise addition.

\subsection{Theoretical Analysis}
\label{subsection.theoretical_analysis}
Here we present analytical comparisons between EL-attention and multi-head attention with/without caching key and value, from the perspective of computational and memory complexity.
Attention related operations are divided into three groups based on arithmetic intensities, see Table~\ref{table.attention.flop}.

\begin{table*}
\caption{\label{table.attention.flop} Computational and memory complexity for three groups of operations in attention. When caching key and value, multi-head attention only calculates key and value for the first time and re-uses them in the following steps. We mark its computational complexity as 0 for simplicity. Our EL-attention does not depend on multi-head key and value from group 1. Assuming sequence length is bigger than the number of heads, it has lower memory complexity for group 3 when using beam search. For notations, $n$ is sequence length, $d_m$ is model dimension, $h$ is number of heads and $x$ is beam size. } 
\centering
\begin{tabular}{l|c|c|c|c|c|c|c}
\hline
 & \multicolumn{3}{c|}{\onemark\: Build Key and Value} & \multicolumn{2}{c|}{\twomark\: Build Query} & \multicolumn{2}{c}{\threemark\: Calculate Attention}  \\
 & \multicolumn{3}{c|}{Compute bound} & \multicolumn{2}{c|}{Compute bound} & \multicolumn{2}{c}{Memory bound}  \\
 \cline{2-8}
 & \multicolumn{2}{c|}{Multi-Head} & EL & {Multi-Head} & EL &{Multi-Head} & EL \\
  &  \multicolumn{2}{c|}{Attention}   &  Attention & {Attention}  &  Attention & {Attention}  &  Attention \\
\hline
\begin{tabular}{@{}l@{}}Cache \\ Key/Value \end{tabular} & \xmark & \cmark & \xmark & \xmark \: or \: \cmark & \xmark & \xmark \: or \: \cmark & \xmark  \\
\hline
\begin{tabular}{@{}l@{}}Compute \\ Complexity\end{tabular} & $\mathcal{O}({nd_m^2})$ & $0$ & $0$ & $\mathcal{O}({d_m^2})$ & $\mathcal{O}({d_m^2})$ & $\mathcal{O}(xnd_m)$ & $\mathcal{O}(hxnd_m)$ \\
\hline
\begin{tabular}{@{}l@{}}Memory \\ Complexity\end{tabular} & $\mathcal{O}(nd_m)$ & $\mathcal{O}(nd_m)$ & $0$ & $\mathcal{O}(d_m)$ & $\mathcal{O}(hd_m)$ & $\mathcal{O}({xnd_m})$ &  $\mathcal{O}({nd_m+hxd_m})$  \\
\hline
\end{tabular}
\end{table*}

The first group of operations is to build multi-head key and value, it has the highest computational complexity $\mathcal{O}(nd_m^2)$ and is bounded by computing, where n is sequence length and $d_m$ is model dimension. Our method EL-attention only depends on previous output. Hence it does not need this part, we mark it as 0 in Table~\ref{table.attention.flop}. To mitigate the computational cost, many sequence-to-sequence libraries \citep{ott-etal-2019-fairseq, wolf-etal-2020-transformers, vaswani-etal-2018-tensor2tensor} support incremental decoding which caches multi-head key and value in each layer. However, a calculation is still needed at the first decode step.
Caching key and value consumes lots of memory, see Table~\ref{table.cache_size}, which prevents using bigger batch size for faster inference speed. It also introduces new maintenance costs. For example, beam candidates' order needs to be changed at each step, and beam search needs to adjust cache order accordingly. Another case happens when part of a batch is finished, the finished sentences need to be removed, it also involves cache adjustment effort.

The second group is to build multi-head query, and project attention results to final output, which is also bounded by computing. EL-attention and multi-head Attention have same computational complexity O($d_m^2$) which is only $1/n$ of the first group. 
It has the shortest execution time among all three groups.

The third group is to calculate attention weights and weighted sum.
This group has $n$ times higher memory complexity compared to group 2, hence bounded by memory. 
Although EL-attention has higher computational complexity than multi-head attention, it has comparable or faster speed due to lower memory complexity.
In beam search or diverse beam search, EL-attention only uses 1/x (beam size) memory movement compared to multi-head attention as presented in \S~\ref{subsubsection.efficient_implementation}.

\section{Experiments}
%This section covers experiments for EL-attention.
First, EL-attention and multi-head attention are compared on synthetic data in \S~\ref{subsection.synthetic}.
Then we integrate EL-attention into existing models and compare performance at model level.
Experiment setup is introduced in \S~\ref{subsection.setup} and main results are presented in \S~\ref{subsection.main_results}.
Speed for various batch sizes is in \S~\ref{subsection.varied_batch_size}. 
Finally, inference accuracy is verified in \S~\ref{subsection.quality}.
Our code is open sourced in \url{https://github.com/microsoft/fastseq}.

\subsection{Analysis on Synthetic Data}  
\label{subsection.synthetic}

\subsubsection{Complexity Requirement} 
\label{subsubsection.complexity}
We collect performance data for each operation in the attention function. When profiling, parameters are set to batch size 32, beam size 4, input sequence length 1024, head number 16, and model dimension 1024.

\begin{figure}[h]
    \centering
	\includegraphics[width = 3.3 in]{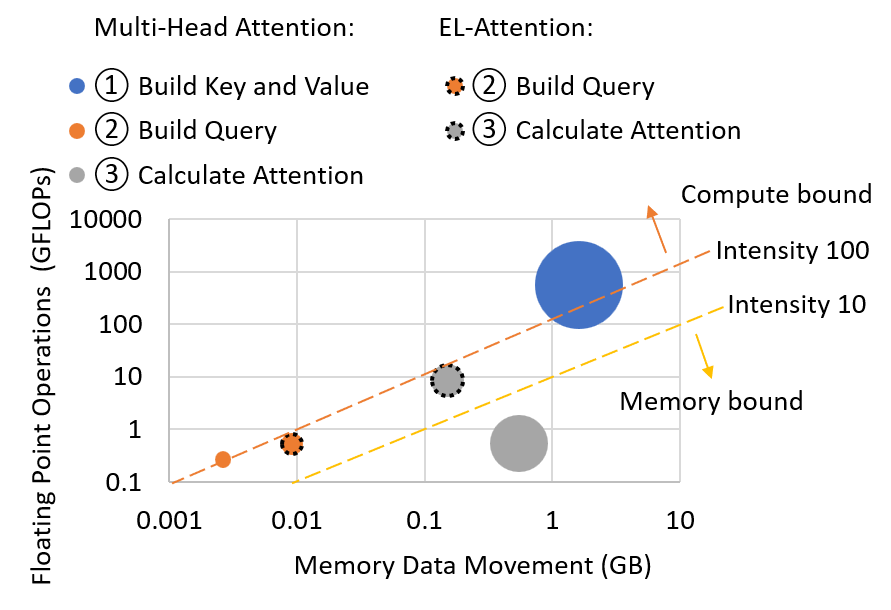}
	\caption{Performance data for operations in attention. X-axis is floating-point operations, y-axis is memory data movement, both are log scaled. Bubble area size represents execution time.
	EL-attention does not have group 1 operations for building key and value. See \S~\ref{subsubsection.complexity} for details. }
	\label{figure.complexity} 
\end{figure}

In Figure~\ref{figure.complexity}, x-axis is memory data movement, y-axis is floating-point operations, both are log scaled, bubble area size represents execution time. 
Our method EL-attention only has two groups of operations, and it is significantly faster than multi-head attention.
In multi-head attention without cache, the first group of operations require more than 500 Giga floating-point operations (GFLOPs) and is undoubtedly bounded by computation.
Arithmetic intensity for group 3 operations is close to 1 in baseline. Recall the threshold for memory-computation even point is from ten to hundred, see \ref{subsecion.speed_limitation}. 
So group 3 of baseline is significantly bounded by memory. Here baseline uses about 0.5 GB memory movement, while our method only costs 0.15 GB.

\subsubsection{Speed Analysis} 
\label{subsubsection.speed_analysis}
\begin{figure*}[h]
    \centering
	\includegraphics[width = 6.5 in]{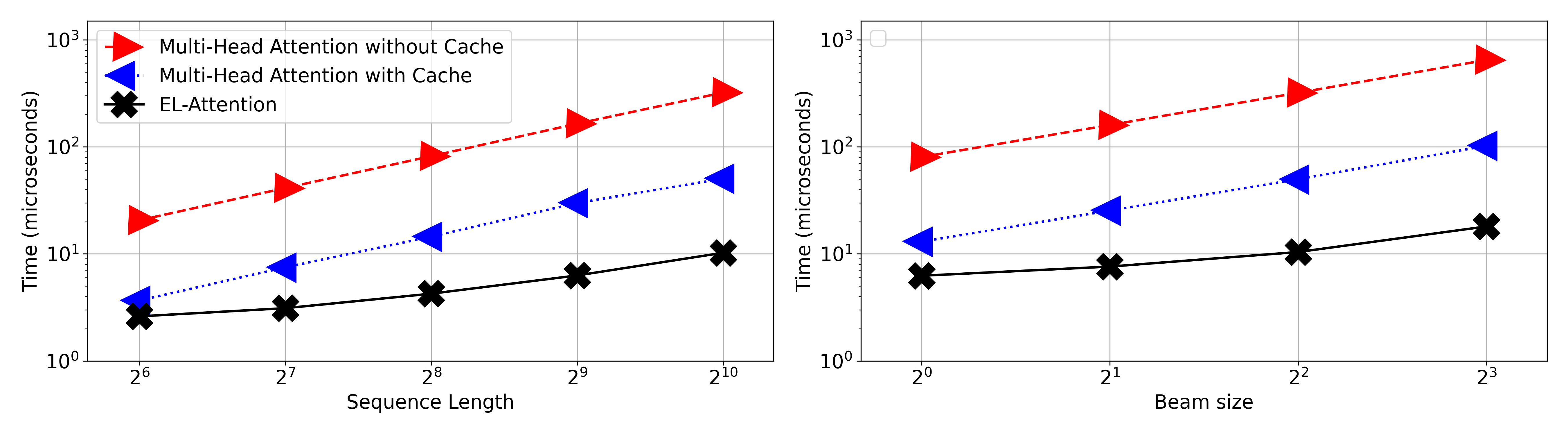}
	\caption{Comparison of attention execution time on various sequence lengths and beam sizes for EL-attention, multi-head attention without cache, and multi-head attention with cache. Axes are log-log scaled. Our method EL-attention has the fastest speed in all settings. Full details of this experiment can be found in \S~\ref{subsubsection.speed_analysis}.}
	\label{figure.input_length}
\end{figure*}
In this subsection, we compare execution time per sample by using synthetic input. 
For studying impact on sequence length $n$, values \{$2^6,2^7,2^8,2^9,2^{10}$\} are used, and beam size is set to 4. For studying impact on beam size $x$, values \{$1,2,4,8$\} are used, and sequence length is set to 1024. 
We scale the batch size inversely with the sequence length and beam size.
When comparing speed, parameter setting is kept the same for three attention methods.
Results are shown in Figure~\ref{figure.input_length}. 

As expected, the execution time for multi-head attention with or without cache increase linearly with respect to sequence length and beam size.
Our method EL-attention is faster than multi-head attention (both with and without cache) for all configurations.
Compared to multi-head attention without cache, EL-attention is up to 36x faster.
Compared to multi-head attention with cache, the speed up ratio grows from 1.4x to 5x when increasing sequence length. Similarly, our method enlarges speed gain from 2x to 5x when increasing beam size from 1 (greedy search) to 8.
EL-attention's execution time increases sub-linearly at the beginning when sequence length and beam size are small.
Starting from certain threshold (around sequence length 256, beam size 4), arithmetic intensity is larger than the even point, it starts increasing linearly due to shift from memory bound to compute bound.

\subsection{Setup} 
\label{subsection.setup}
In this section, we introduce datasets and models used for studying inference speed in this work. 
We conduct experiments on a NVIDIA Tesla V100 PCIe 16GB. We choose the maximum batch size that can fit in GPU memory.

\subsubsection{Datasets} 
Two summarization datasets and a question generation dataset are used as benchmarks. 

\noindent\textbf{SQuAD 1.1} \citep{rajpurkar-etal-2016-squad} contains over 100K questions in 536 Wikipedia articles. The version redistributed by GLGE~\cite{liu2020glge} has 75722/10570/11877 samples in training/validation/test set. The average input and output length are 149.4 and 11.5 respectively.

\noindent\textbf{XSum} \citep{narayan-etal-2018-dont} 
consists online articles from BBC. There are 204017/11327/11333 samples in training/validation/test set.
The average input and output length are 358.5 and 21.1 respectively.

\noindent\textbf{CNN/DailyMail} \cite{NIPS2015_afdec700} contains articles from CNN and Daily Mail newspapers. 
There are 287113/13368/11490 samples in training/validation/test set. 
The average input and output length are 822.3 and 57.9.

\subsubsection{Models} 
EL-attention can be applied to a wide range of multi-head attention based models. Here we select three popular models (Transformer, BART, and GPT-2) as representatives. They cover two widely used architectures: encoder-decoder and decoder only. Experiments load their existing model checkpoints without extra training effort. 
 
\noindent\textbf{Transformer} \citep{NIPS2017_3f5ee243} is a widely studied encoder-decoder model with attention function. We use checkpoint shared by GLGE~\citep{liu2020glge} which is trained from scratch on SQuAD 1.1.

\noindent\textbf{BART} ~\citep{lewis-etal-2020-bart} is another popular encoder-decoder model which is pretrained via denoising. We use two official released checkpoints which are finetuned on CNN/DailyMail and XSum respectively.

\noindent\textbf{GPT-2} ~\citep{radford2019language} is a decoder only model, we load its released pretrain checkpoint and do inference on summarization task by following their paper. 

For Transformer model and BART model, we use implementations in FairSeq~\citep{ott-etal-2019-fairseq} v0.9.0; for GPT-2 model, we use Huggingface Transformers~\citep{wolf-etal-2020-transformers} v3.0.2.
Based on these implementations, multi-head attention is replaced by our EL-attention.

\subsubsection{Inference Parameters}
First, we list parameters for beam search. In SQuAD 1.1 task, we set length penalty to 1.0, max input length is 512, and beam size is 4.
In XSum task, we use parameters listed in BART\footnote{https://github.com/pytorch/fairseq/blob/master/examples/bart}, with length penalty 1.0, max input length 1024, max output length  60, min output length 10, and beam size 6. 
In CNN/DailyMail task, we conduct experiments for both BART and GPT-2. For BART model, we follow their parameters, with length penalty 2.0, max input length 1024, max output length 140, min output length 55, and beam size 4. For GPT-2 model, following their paper, we directly use pretrained model checkpoint to generate summarization, max input length is 512, max output length is set to 200. 

In diverse beam search, we set diverse beam group same as beam size, and diverse beam strength is set to 0.2.

When switching from beam search to greedy search, we change the beam size to 1 and keep all other parameters unchanged.

\begin{table*}
\centering
\caption{\label{table.main_results} Inference speed (samples/second) comparison across decoding methods, models, tasks, and computation precision. Speed up ratio is calculated within the same precision. Single precision marked as fp32 and half precision marked as fp16. See details in \S~\ref{subsection.main_results}.}
\begin{tabular}{lllccc}
\hline 
{Model} & {Parameter} & {Task} &  {{Multi-Head Attention}} &  {{EL-Attention}} &  {{Speed Up Ratio}}  \\ 
& {Number} &  &  {fp16 (fp32)} & {fp16 (fp32)} & {fp16 (fp32)}  \\ \hline
\multicolumn{6}{c}{{Beam Search}} \\
\hline 
Transformer & 270M & SQuAD 1.1 & 170.9 (86.6) & 458.1 (173.6) & \textbf{2.7x} (2.0x)  \\
BART$_{\textrm{large}}$ & 400M & XSum & 14.7 (6.8) & 69.4 (26.3)  & \textbf{4.7x} (3.9x)  \\
BART$_{\textrm{large}}$  & 400M & CNN/DailyMail  & 5.7 (3.4) & 28.6 (12.2)  & \textbf{5.0x} (3.6x) \\
GPT-2$_{\textrm{small}}$ & 117M & CNN/DailyMail & 2.1 (1.5) & 3.8 (2.5) & \textbf{1.8x} (1.7x)  \\
GPT-2$_{\textrm{medium}}$ & 345M & CNN/DailyMail & 0.9 (0.6) & 2.0 (1.1) & \textbf{2.2x} (1.8x)  \\
\hline 
\multicolumn{6}{c}{{Diverse Beam Search}} \\
\hline 
Transformer & 270M & SQuAD 1.1 & 162.3 (82.1) & 454.1 (171.8) & \textbf{2.8x} (2.1x)  \\
BART$_{\textrm{large}}$  & 400M & XSum & 15.8 (7.2) & 71.9 (27.5) &  \textbf{4.6x} (3.8x)   \\
BART$_{\textrm{large}}$  & 400M & CNN/DailyMail & 5.4 (3.2) & 28.5 (12.0) & \textbf{5.3x} (3.8x) \\
\hline 
\multicolumn{6}{c}{{Greedy Search}} \\
\hline 
Transformer & 270M & SQuAD 1.1 & 436.4 (190.3)  & 699.7 (260.3)  & \textbf{1.6x} (1.4x)  \\
BART$_{\textrm{large}}$  & 400M & XSum & 42.6 (15.0) & 107.8 (44.9)  & \textbf{2.5x} (3.0x)  \\
BART$_{\textrm{large}}$  & 400M & CNN/DailyMail & 13.0 (7.0) & 40.0 (19.5) & \textbf{3.1x} (2.8x) \\
GPT-2$_{\textrm{small}}$ & 117M & CNN/DailyMail & 14.7 (9.0) & 26.2 (13.6) & \textbf{1.8x} (1.5x) \\
GPT-2$_{\textrm{medium}}$ & 345M & CNN/DailyMail & 5.9 (3.6) & 10.4 (5.9) & \textbf{1.8x} (1.6x) \\
\hline
\end{tabular}
\end{table*}

\subsection{Main Results}  
\label{subsection.main_results}
We present inference speed for three model types (total five models), three tasks, three decoding methods (beam search, diverse beam search and greedy search) and two computation precision settings (fp16 and fp32) in Table~\ref{table.main_results}.
Compared with the baseline, EL-attention achieves 1.6x to 5.3x speedup for all models on these tasks and decoding methods.

In beam search, the speed-up ratio is between 1.8x and 5.0x for half precision and 1.7x to 3.9x for single precision. 
For example, EL-attention has 5x speed-up for BART$_\textrm{large}$ model when inference on the CNN/DailyMail. 
EL-attention shows more than twice speed up for Transformer model on SQuAD 1.1. 
For GPT-2 model, speed is 1.8x and 2.2x for small and medium model size, respectively\footnote{The absolute speed for GPT-2 model are smaller due to a bottleneck operation in Huggingface Transformers v3.0.2, which searches for finished beams sequentially.}.

In diverse beam search, there is a similar speedup ratio as that in beam search\footnote{Diverse beam search experiments for GPT-2 model are skipped because we did not find this feature in Huggingface Transformers v3.0.2.}.
Transformer model on SQuAD 1.1 dataset is accelerated to 2.8x, BART model on XSum and CNN/DailyMail has speed up 4.6x and 5.3x, respectively. 

In greedy search, EL-attention achieves speedup from 1.6x to 3.1x for half precision. 
For example, speedup of BART model is 3.1x on summarization tasks CNN/DailyMail. GPT-2 model on this task shows 1.8x speed up. 

In general, EL-attention has more significant speed gain for longer input and larger model size.
The speedup ratio is higher when using half precision than single precision, due to half precision has higher arithmetic intensity threshold that balances memory bound and compute bound.

\subsection{Speed on Various Batch Sizes}  
\label{subsection.varied_batch_size}
In this section, We study inference speed on varied batch sizes and present their cache size differences. 
We include another baseline multi-head attention without cache here.

In Table~\ref{table.batch_size}, compared with multi-head attention without cache, when using the same batch size 32, speed is 3x by using cache, speed is 4.2x by using EL-attention. 
Our method can further enlarge speed gain to 15.1x when batch size grows to 320.
Here are two explanations why bigger batch increases speed: 1) higher arithmetic intensity from more calculations per memory movement; 2) longer execution time per instruction, which mitigates latency between CPU and GPU communication.
As expected, among the three methods, multi-head attention with cache consumes the most memory, therefore, supports the smallest batch size. 
While EL-attention can support batch size up to 320, multi-head attention without cache will be OOM at batch size 128, because it re-computes states for all previously generated tokens at each step, which consumes more run-time memory.
Multi-head attention without cache has no speed gain by increasing batch size since it already has high arithmetic intensity and long execution time per instruction even in small batch size. However, most of these calculations are duplicated efforts.

\begin{table}
\centering
\caption{\label{table.batch_size} Inference speed (samples/second) on different batch sizes. OOM means out of memory. All speedup ratios are compared to the same cell value of no cache and batch size 32.}
\begin{tabular}{l|c|c|c}
\hline 
  & \multicolumn{2}{c|}{{Multi-Head Attention}}  & EL-Attention \\ 
 Batch size & No Cache & Has Cache  &  \\ 
\hline 
 32 & 1.9 (1x) & 5.7 (3x) & 8.0 (4.2x) \\
 64 & 1.9 (1x) & OOM & 12.6 (6.6x) \\
 128 & OOM & OOM & 21.3 (11.2x) \\
 320 & OOM & OOM & 28.6 (15.1x) \\
\hline
\end{tabular}
\end{table}

We compare cache sizes required for storing input related model states between multi-head attention and EL-attention in Table~\ref{table.cache_size}. 
Our method only stores encoder output, so the memory size is 96x smaller compared to multi-head attention.
Refer to Figure~\ref{figure.incremental} for the relationship between encoder output and key-value pairs. 
These numbers are calculated based on BART$_{\textrm{large}}$ and half precision.

\begin{table}
\centering
\caption{\label{table.cache_size} Comparison of memory sizes used for storing input related model states, see \S~\ref{subsection.varied_batch_size} for detail. }
\begin{tabular}{lcc}
\hline 
 Sequence & Multi-Head   & EL \\ 
 Length & Attention & Attention   \\ 
\hline 
\multicolumn{3}{c}{{Batch size 32}} \\ 
\hline 
 256 & 1.5 GB & 0.02 GB \\
 1024 & 6 GB & 0.06 GB \\
 \hline
 \multicolumn{3}{c}{{Batch size 64}} \\ 
\hline 
 256 & 3 GB & 0.03 GB \\
 1024 & 12 GB & 0.13 GB \\
 \hline
 \multicolumn{3}{c}{{Batch size 320}} \\ 
\hline 
 256 & 15 GB & 0.15 GB \\
 1024 & 60 GB & 0.63 GB \\
\hline
\end{tabular}
\end{table}

\subsection{Accuracy Verification}  
\label{subsection.quality}
%In this section, we verify the generation result of EL-attention on CNN/DailyMail test set.
%All parameters are kept the same as those used in BART paper, with beam size 4, length penalty 2.0, min and max output length are 55 and 140 respectively, and no repeat ngram is set to 3. 
At first, we direct load the BART released checkpoint for inference, then calculate the ROUGE score. Our reproduced results are similar to the numbers reported in their paper.
Then we replace multi-head attention with EL-attention, there is no result change when using single precision (fp32). 
Compared result between half precision (fp16) and fp32, there are slight differences in both multi-head attention and EL-attention due to low precision. But the rouge score differences are minimal.
We report all the ROUGE scores in Table~\ref{table.quality}. 
%It proves EL-attention can safely replace multi-head attention for both fp32 and fp16 computations.
\begin{table}
\centering
\caption{\label{table.quality}  Evaluate EL-attention's impact on generation quality using CNN/DailyMail test set. }
\setlength\tabcolsep{4pt}
\begin{tabular}{l@{\hskip2pt}ccc}
\hline 
 Model  &  ROUGE-1 & ROUGE-2 & ROUGE-L  \\ 
\hline 
BART$_{\textrm{large}}$   & 44.16 & 21.28 & 40.90  \\
\hline
Our reproduce (fp32) & 44.21 & 21.20 & 41.03  \\
+EL-attention (fp32) & 44.21 & 21.20 & 41.03  \\
\hline
Our reproduce (fp16)  & 44.22 & 21.20 & 41.04  \\
+EL-attention (fp16) & 44.22 & 21.21 & 41.05  \\
\hline
\end{tabular}
\end{table}

\section{Related Work}
\subsection{Transformer Speed up}  
Many works focus on improving inference speed for Transformer~\cite{NIPS2017_3f5ee243} and variant models. 
1) Reducing complexity on sequence length.
PoWER-BERT~\citep{goyal2020power} studies progressive word-vector elimination, Linformer~\citep{wang2020linformer} proposals attention with linear complexity, Reformer~\citep{Kitaev2020Reformer:} reduces complexity by locality-sensitive hash, BigBird~\citep{NEURIPS2020_c8512d14} and LongFormer~\citep{beltagy2020longformer} proposes sparse attention with global tokens. 
Linear Transformers~\citep{pmlr-v119-katharopoulos20a} only stores accumulated states instead of maintaining every representation.
Sparse Sinkhorn Attention~\citep{tay2020sparse} reduces memory complexity based on differentiable sorting. 
2) Reducing model size.
Fixed Multi-Head Attention~\citep{pmlr-v119-bhojanapalli20a} studies choosing $d_k$ based on the input sequence length. 
One Write-Head~\citep{shazeer2019fast} shares one head of key and value for multi-head query and achieves speed up with minor quality degradation. 
LayerDrop~\citep{fan2019reducing} enables efficient layer pruning at inference stage. 
3) Non-autoregressive generation. 
\citet{gu2018nonautoregressive, lee-etal-2018-deterministic, qi2020bang} speed up inference by predicting all tokens in single step instead of step-by-step generation.

\subsection{Roofline Performance Analysis}
\label{subsection.roofline_analysis}
The Roofline model~\citep{williams2009roofline} provides an intuitive and insightful approach to identifying performance bottleneck. The attainable floating-point performance is affected by peak floating-point performance, peak memory bandwidth, and arithmetic intensity. 
\citet{yang2020hierarchical} constructs a hierarchical Roofline on NVIDIA GPU and extends it to support half precision and Tensor Cores, this hierarchical Roofline incorporates L1, L2, device memory, and system memory bandwidths. 
\citet{wang2019benchmarking} studies this problem across platforms on TPU, GPU, and CPU.
\citet{wang2020hierarchical} presents a practical methodology for collecting performance data to conduct hierarchical Roofline analysis on NVIDIA GPU. 

\section{Conclusion}
In this work, we present EL-attention, a technology that significantly reduces memory cost and increases inference speed. 
%By exploiting the associative property of matrix multiplication, we prove EL-attention can produce the same result as vanilla multi-head attention.
EL-attention can be directly applied to existing model checkpoints without accuracy loss. Because of the massive memory savings, it might be especially helpful for efficient inference on memory limited devices, like mobile and IoT devices. We leave them as future work.

\newpage

\bibliography{paper}
\bibliographystyle{icml2021}

\newpage \:
\newpage
\appendix

\section{Pseudocode}
To better understand differences between EL-attention and multi-head attention in generation, we list their pseudocode for the comparison. For simplicity, bias term and some non-essential operations are omitted.

Algorithm~\ref{alg:generation_process} shows the generation process for a typical encoder-decoder model. First, input is  encoded by model, then decoder starts repetitively executing, one token is generated per step. The high level logic is the same for both attention methods, differences are in the cache and  computation. 

The generation process with multi-head attention is listed in Algorithm~\ref{alg:multi_head_attention}. First, the encoder output is \colorbox{red!30}{\textrm{repeated}(h)} beam size times to match query's first dimension. Second, the key and value \colorbox{yellow!30}{cache are built for every layer}. So its cache size is linear to  number of the decoder layer $L$, beam size $x$ and  batch size $b$. 

EL-attention (See Algorithm~\ref{alg:el_attention}) does not need to repeat the encoder output and get rid of the per layer cache for storing key and value. 
To achieve these benefits, EL-attention shifts some computation on key and value to the query side. Due to the length of query is always one which is much shorter than the length of key and value in most cases, it reduces the computations significantly.
%EL-attention has two extra operations compared to multi-head attention after the first generation step. 
%But these new operations are unrelated to key and value, which are much faster than the existing operations. 

EL-attention can achieve faster speed due to time saved from \colorbox{blue!30}{reorder\_cache()},  two \colorbox{orange!30}{\textrm{torch.bmm}} operations, and the ability to use much larger batch size.
%This big cache limited the maximum batch size and slow down inference in the step of \colorbox{blue!30}{reorder\_cache()}. And in the two \colorbox{orange!30}{\textrm{torch.bmm}}, the k and v size are linear to $x$. 

\begin{figure*}
\begin{minipage}{1.0\textwidth}
\begin{algorithm}[H]
\hsize=\textwidth % <--------- THE HACK!
   \caption{Generation Process}
   \label{alg:generation_process}
\begin{algorithmic}
   \STATE {\bfseries Input:} data $src\_tokens$, beam size $x$
   \STATE {\bfseries Output:} $tokens$
   \STATE $encoder\_outs = \textrm{forward\_encoder}(src\_tokens)$
   \STATE Initialize $previous\_output[:] = \textrm{BOS}$
   \STATE Initialize $tokens = array$
   \FOR{$t=0$ {\bfseries to} $T$}
   \STATE $logits = \textrm{forward\_decoder}(previous\_output, encoder\_outs)$
   \STATE $previous\_output, order\_index = \textrm{sample}(logit, x)$
   \STATE $tokens = \textrm{reorder}(tokens, order\_index)$
   \STATE $tokens[t, :] = previous\_output$
   \ENDFOR
\end{algorithmic}
\end{algorithm}
\end{minipage}
\begin{minipage}[t]{0.5\textwidth}
\begin{algorithm}[H]
\hsize=\textwidth % <--------- THE HACK!
   \caption{forward\_decoder with multi-head attention}
   \label{alg:multi_head_attention}
\begin{algorithmic}
   \FUNCTION{forward\_decoder(previous\_output, h)}
   \IF{$cache$ is None}
   \STATE \colorbox{red!30}{$h = \textrm{repeat}(h)$} \COMMENT{repeat beam size times}
   \ELSE
   \STATE \colorbox{blue!30}{reorder\_cache()} \COMMENT{Cache size: O(2BLSD), where B is beam size, L is decoder layer, S is sequence length, D is model dimension.}
   \ENDIF
   \STATE $x = \textrm{embedding}(previous\_output)$
   \FOR{$i=0$ {\bfseries to} layers $L$}
   \STATE $x = \textrm{self\_attention}(x)$
   \STATE $x = \textrm{encoder\_decoder\_attention}(x, h, h)$
   \STATE $x = \textrm{mlp}(x)$
   \ENDFOR
   \STATE {\bfseries return} $\textrm{predict\_on\_vocab}(x, unembedding\_weight)$
   \ENDFUNCTION
   \STATE 
   \FUNCTION{encoder\_decoder\_attention(query, key, value)}
   \STATE \COMMENT{$query \in [bx, 1, d_m]$}
   \IF{$cache[i,k]$ is None}
   \STATE \colorbox{yellow!30}{$cache[i,k] = \textrm{reshape}(\textrm{torch.mm}(key, W_{k}^{i}))$}
   \STATE \colorbox{yellow!30}{$cache[i,v] = \textrm{reshape}(\textrm{torch.mm}(value, W_{v}^{i}))$}
   \ENDIF
   \STATE $k = cache[i,k]$ \COMMENT{$k \in [bxh, d_k, n]$}
   \STATE $v = cache[i,v]$ \COMMENT{$v \in [bxh, n, d_k]$}
   \STATE \COMMENT{$q \in [bxh, 1, d_k]$}
   \STATE $q = \textrm{reshape}(\textrm{torch.mm}(query, W_{q}^{i}))$
   \STATE \colorbox{orange!30}{$weights = \textrm{torch.bmm}(q, k)$}
   \STATE $prob = \textrm{softmax}(weights)$
   \STATE \colorbox{orange!30}{$attn = \textrm{torch.bmm}(prob, v)$}
   \STATE $attn = \textrm{torch.mm}(attn, W_{o}^{i})$
   \ENDFUNCTION
\end{algorithmic}
\end{algorithm}
\end{minipage}\hfill
\begin{minipage}[t]{0.5\textwidth}
\begin{algorithm}[H]
\hsize=\textwidth % <--------- THE HACK!
   \caption{forward\_decoder with EL-attention}
   \label{alg:el_attention}
\begin{algorithmic}
   \FUNCTION{forward\_decoder(previous\_output, h)}
   \IF{$cache$ is not None}
   \STATE \colorbox{blue!30}{reorder\_cache()} \COMMENT{Cache size: O(SD), where S is sequence length, D is model dimension. Which is 2BL times less. }
   \ENDIF
   \STATE $k = \textrm{reshape}(h)$ \COMMENT{$k \in [b, d_m, n]$}
   \STATE $v = \textrm{reshape}(h)$ \COMMENT{$v \in [b, n, d_m]$}
   \STATE $x = \textrm{embedding}(previous\_output)$
   \FOR{$i=0$ {\bfseries to} layers $L$}
   \STATE $x = \textrm{self\_attention}(x)$
   \STATE $x = \textrm{encoder\_decoder\_attention}(x, k, v)$
   \STATE $x = \textrm{mlp}(x)$
   \ENDFOR
   \STATE {\bfseries return} $\textrm{predict\_on\_vocab}(x, unembedding\_weight)$
   \ENDFUNCTION
   \STATE
   \FUNCTION{encoder\_decoder\_attention(query, k, v)}
   \STATE \COMMENT{$query \in [bx, 1, d_m]$}
   \STATE \COMMENT{No heavy op for building multi-head key/value.}
   \STATE \COMMENT{Encoder output is directly used as key and value, and shared among all layers.}
   \STATE $q = \textrm{reshape}(\textrm{torch.mm}(query, W_{q}^{i}))$  \COMMENT{$q \in [bx, h, d_k]$}
   \STATE \COMMENT{$W_{k}^{i} \in [h, d_k, d_m]$}
   \STATE \colorbox{yellow!30}{$q = \textrm{reshape}(\textrm{torch.bmm}(q, W_{k}^{i}))$}  \COMMENT{$q \in [b, hx, d_m]$}
   \STATE \colorbox{orange!30}{$weights = \textrm{torch.bmm}(q, k)$}
   \STATE $prob = \textrm{softmax}(weights)$
   \STATE \colorbox{orange!30}{$attn = \textrm{torch.bmm}(prob, v)$} \COMMENT{$attn \in [b, hx, d_m]$}
   \STATE \COMMENT{$W_{v}^{i} \in [h, d_m, d_k]$}
   \STATE \colorbox{yellow!30}{$attn = \textrm{reshape}(\textrm{torch.bmm}(attn, W_{v}^{i}))$}
   \STATE $attn = \textrm{torch.mm}(attn, W_{o}^{i})$
   \ENDFUNCTION
\end{algorithmic}
\end{algorithm}
\end{minipage}
\end{figure*}

\section{Proof for EL-Attention}
In this section, we will present the proof that  EL-attention can have the same result as multi-head attention via the choice of FFN functions. And again, there is no explicit conversion on key and value.

Recall that, by expanding Equation 1 and 2 from the paper, multi-head attention can be formulated as:
\begin{equation} \tag{6}
\begin{aligned}
&\textrm{MultiHead}(Q, K, V) = \sum_{i=1}^{h} \overbrace{\textrm{softmax}(\frac{Q_iK_i^T}{\sqrt{d_k}})}^{\textrm{Prob}_i}V_iW_i^O \\
&\textrm{where}\: Q_i=QW_i^Q+b_i^Q, \: K_i=KW_i^K+b_i^K, \\
&\:\:\:\:\:\:\:\:\:\:\:\: V_i=VW_i^V+b_i^V, \: \textrm{and}\: H = K = V
\end{aligned}
\label{appendix.eq.multi-head-attention}
\end{equation}
Here, Q $\in$ $\mathbb{R}^{1\times d_m}$, H $\in$ $\mathbb{R}^{n\times d_m}$, 
$W_i^Q$, $W_i^K$, $W_i^V$ $\in$ $\mathbb{R}^{d_m\times d_k}$ and $W_i^O$ $\in$ $\mathbb{R}^{d_k\times d_m}$.
We include bias term in this proof, it is omitted in previous equations for simplification. 

The multiplication of single head query and single head key can be replaced by the multiplication of expanded query and original key, derived as:
\begin{equation} \tag{7}
\begin{aligned}
Q_iK_i^T &= (QW_i^Q+b_i^Q)(KW_i^K+b_i^K)^T \\
&= (QW_i^Q+b_i^Q)((KW_i)^T \\
&\:\:\:\:\:+ (QW_i^Q+b_i^Q)(b_i^K)^T \\
&= \textrm{FFN}_i^Q(Q)K^T + Q_i(b_i^K)^T \\
\textrm{where}\: &\textrm{FFN}_i^Q(Q)=(QW_i^Q+b_i^Q)(W_i^{K})^T \\
&\textrm{and}\: Q_i=QW_i^Q+b_i^Q \\
\end{aligned}
\label{appendix.eq.qk}
\end{equation}

Below is the EL-attention conversion from single head attention result to final output in multi-head attention:
\begin{equation} \tag{8}
\begin{aligned}
\textrm{Prob}_i\cdot V_i\cdot W_i^O &= \textrm{Prob}_i(VW_i^V+b_i^V)W_i^O \\
&= \textrm{Prob}_i(VW_i^V)W_i^O \\
&\:\:\:\:\:+ \textrm{Prob}_i\cdot \textrm{Repeat}(b_i^V)\cdot W_i^O \\
&= \textrm{FFN}_i^O(X) + b_i^VW_i^O \\
\textrm{where}\: &\textrm{FFN}_i^O(X)=XW_i^VW_i^O \\
&\textrm{and}\: X=\textrm{Prob}_i\cdot V \\
&\textrm{and}\: \textrm{Repeat}(b_i^V) \: \textrm{is broadcasting dim} \\
\end{aligned}
\label{appendix.eq.out}
\end{equation}

To ensure the equivalence to multi-head attention, we adjust EL-attention as:
\begin{equation} \tag{9}
\begin{aligned}
&\textrm{EL}(Q, K, V) = \sum_{i=1}^{h} \textrm{FFN}_i^O(\textrm{Prob}_i\cdot V) + \sum_{i=1}^{h} b_i^VW_i^O \\
&\textrm{where}\: \textrm{Prob}_i=\textrm{softmax}(\frac{\textrm{FFN}_i^Q(Q)K^T + Q_i(b_i^K)^T}{\sqrt{d_k}}) \\
&\:\:\:\:\:\:\:\:\:\:\:\: \textrm{and}\: H = K = V
\end{aligned}
\label{appendix.eq.el-attention}
\end{equation}

By leveraging the associative property of matrix multiplication, Equation~\ref{appendix.eq.multi-head-attention} and Equation~\ref{appendix.eq.el-attention} are interchangeable. 

Please note that some bias terms can be omitted when training a new model. 
Like the bias term $b_i^K$ that adding the same value for all attention positions, and $b_i^V$ that contributing constant information to the output, it is independent of query/key/value and the sum of all elements in Prob$_i$'s last dimension is always one.

\end{document}